\documentclass[conference]{IEEEtran}
\IEEEoverridecommandlockouts
\usepackage{cite}
\usepackage{amsmath,amssymb,amsfonts}
\usepackage{algorithmic}
\usepackage{graphicx}
\usepackage{textcomp}
\usepackage{xcolor}
\usepackage{float}
\usepackage{array}
\usepackage{hyperref}

\def\BibTeX{{\rm B\kern-.05em{\sc i\kern-.025em b}\kern-.08em
    T\kern-.1667em\lower.7ex\hbox{E}\kern-.125emX}}
\begin{document}

\title{Predicting Chaotic System Behavior using Machine Learning Techniques\\
\thanks{\textsuperscript{1} School of Electrical and Computer
Engineering, Georgia Institute of Technology, Atlanta, GA, 30308, USA. \texttt{\{hrao43, yzhao654\}@gatech.edu}\par
\textsuperscript{2} School of Electrical and Automation Engineering, East China Jiaotong University, Nanchang 330013, China \texttt{laiqiang87@126.com}\par
}
}

\author{\IEEEauthorblockN{Huaiyuan Rao\textsuperscript{1}, Yichen Zhao\textsuperscript{1}, Qiang Lai\textsuperscript{2}}}
\maketitle

\begin{abstract}
Recently, machine learning techniques, particularly deep learning, have demonstrated superior performance over traditional time series forecasting methods across various applications, including both single-variable and multi-variable predictions. This study aims to investigate the capability of i) Next Generation Reservoir Computing (NG-RC) ii) Reservoir Computing (RC) iii) Long short-term Memory (LSTM) for predicting chaotic system behavior, and to compare their performance in terms of accuracy, efficiency, and robustness. These methods are applied to predict time series obtained from four representative chaotic systems including Lorenz, Rössler, Chen, Qi systems. In conclusion, we found that NG-RC is more computationally efficient and offers greater potential for predicting chaotic system behavior.
\end{abstract}

\section{Introduction}
Time series data have attracted significant attention across various fields in the natural and social sciences because of their potential applications. The analysis and prediction of time series data have been the focus of extensive research over the past few decades \cite{farmer1987predicting, muller1997predicting, sapankevych2009time, weigend2018time, torres2021deep}. Chaotic time series are among the most complex because even small perturbation in initial values can lead to significant variations in their behaviors. Due to their sensitivity to initial conditions, it is a challenging task to predict chaotic time behaviors.\par

A practical solution is to develop models that forecast the behavior of chaotic time series. Recently, data-driven approaches such as machine learning (ML) have shown its efficiency on chaotic time series forecasting. For instance, recurrent neural networks (RNNs) are widely used across various fields in engineering and science for learning sequential tasks or modeling and predicting time series \cite{ahmed2010empirical, shi2017deep, tsai2018air}. Yet, they struggle with long-term temporal dependencies, slow and subtle changes, or significantly varying time scales because their loss gradients backpropagated in time tend to saturate or diverge during training \cite{mikhaeil2022difficulty}. One solution to this is based on specifically designed RNN architectures with gating mechanisms, such as long short-term memory (LSTM) \cite{hochreiter1997long}, which enables states from earlier time steps to more easily influence activities occurring much later by using a protected memory buffer. \par

Reservoir computing (RC) has shown the potential of achieving higher-precision prediction of chaotic time series \cite{RC1, RC2, RC3}. The core idea is to utilize dynamical systems as reservoirs (nonlinear generalizations of standard bases) to adaptively learn spatiotemporal features and hidden patterns in complex time series \cite{yan2024emerging}. A RC model is based on a recurrent artificial neural network with a pool of interconnected neurons. To avoid the vanishing gradient problem during training, the RC paradigm randomly assigns the input-layer and reservoir link weights. Unlike other machine learning methods which are computational costly, only the weights of the output links are trained via a regularized linear least-squares optimization procedure. 
Due to its simple structure, this method has been used for multi-step-ahead predictions of nonlinear time series and for modeling chaotic dynamical systems with low computational cost \cite{bo2020asynchronously, griffith2019forecasting, pathak2018model}. 

The RCs have been developed from the original Echo-state network (ESN)-based to nonlinear vector autoregression (NVAR), which also called NG-RC \cite{gauthier2021next}. An NVAR machine is created where the feature vector is composed of time-delayed observations of the dynamical system, along with nonlinear functions of these observations. It requires no random matrices, fewer metaparameters, and provides interpretable results which reflects the nature of the nonlinear model. In addition, it is $\sim33-162$ times less costly to simulate than a typical already efficient traditional RC \cite{gauthier2021next}. Due to these advantages, it has the ability to accurately forecast short term dynamics, reproduce long-term chaotic system and infer the behavior of unseen data of a dynamical system. \par

The main contribution is threefold. Firstly, We compared the capability of three ML methods for forecasting chaotic time series obtained from four representative dynamical systems. Then, we further compare RC and NG-RC's ability on predicting chaotic time series using less observable dataset and extend its predicting time horizon. At last, We test RC and NG-RC's robustness on predicting chaotic time series derived from noisy system ODEs .\par 

Our paper is organized as follows. In Section II we formulate the bench-marking prediction problems. Section III presents a summary of the modeling approaches used for forecasting chaotic time series, which provides detailed information about the implementation of each model and the evaluation metrics employed in the study. In Section IV we apply these methods to finish different tasks corresponding to each problem and discuss each methods' pros and cons. In Section V, we present our conclusions.
\newpage

\section{Problem Formulation}
A general dynamical process can be described by the following ordinary differential equation (ODE):
\begin{equation}
    \dot{x}(t) = f(x(t))
\end{equation}
where $f(\cdot)$ is system dynamics, $x(t) \in \mathbb{R}^n$ is the state vector and $t$ denotes time. \\
\textbf{Problem 1 (Multi-steps Forecasting Chaotic Time Series)} 
Suppose in a finite time horizon $t\in [0, T]$, the state vector $x(t)$ is only observable in a limited time period $[T_1, T_2]$. where $T_1\geq 0, T_2 \leq T$. Meanwhile we have corresponding time sequence data of state $x(T_1), x(T_1+t_0), x(T_1+2t_0), \cdots, x(T_2)$ with time step $t_0$. Our aim is using these time-limited observable state data (with noise in most cases) to predict the real system state in time horizon $t\in[T_2, T]$.\\
\textbf{Problem 2 (Predicting Noisy Chaotic System Behavior)}
Dynamical system process is easily affected by unknown noise in real world. Usually the observed state of a system is a combination of real system state $x(t) \in \mathbb{R}^n$ and noise $\mathcal{E} \in \mathbb{R}^n$. Using noisy input data represents a more realistic implementation in practical applications for system behavior predictions. In short, predicting a noise system can also test algorithm's robustness when encounter some unknown perturbations (e.g. friction, wind).\\

\section{Models and Methods}
\subsection{Long Short-Term Memory Neural Network (LSTM)}
\begin{figure}[t]
    \centering
    \includegraphics[width=1\linewidth]{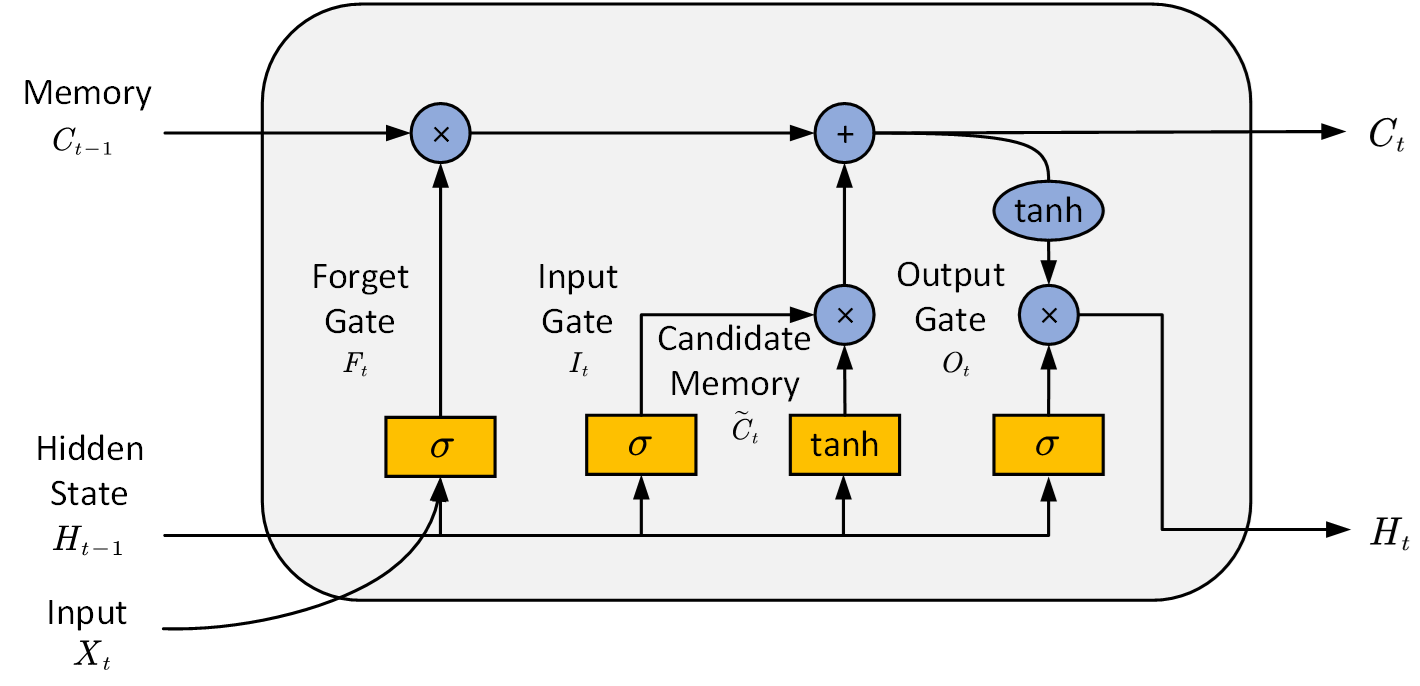}
    \caption{The LSTM cell's structure}
    \label{fig:LSTM}
\end{figure}

A LSTM recurrent neural network processes a variable-length sequence $x=(x_1, x_2, \cdots, x_n)$ by incrementally adding new content into a single memory slot with gates controlling the extent to which new content should be memorized, old content should be erased, and current content should be exposed. The compact forms of the equations for the forward pass of an LSTM cell with a forget gate are
\begin{equation}
\begin{aligned}
f_t &= \sigma_g(W_f x_t + U_f h_{t-1} + b_f) \\
i_t &= \sigma_g(W_i x_t + U_i h_{t-1} + b_i) \\
o_t &= \sigma_g(W_o x_t + U_o h_{t-1} + b_o) \\
\tilde{c}_t &= \sigma_c(W_c x_t + U_c h_{t-1} + b_c) \\
c_t &= f_t \odot c_{t-1} + i_t \odot \tilde{c}_t \\
h_t &= o_t \odot \sigma_h(c_t)
\end{aligned}
\end{equation}
where $x_t \in \mathbb{R}^d$ is the input vector to the LSTM unit, $f_t, i_t, o_t$ are gate activation vector. $h_t, \tilde{c}_t$ are output vector and input activation vector, respectively. $c_t \in \mathbb{R}^h $ is cell state vector. $W \in \mathbb{R}^{h\times d}, U \in \mathbb{R}^{h\times h}$ and $b\in \mathbb{R}^h$ are weight matrices and bias vector parameters which need to be learned during training (Fig. \ref{fig:LSTM}). \par
LSTM is a type of RNN which is useful to handle sequence prediction problems with input data that has long-range temporal dependencies. It is designed to retain information from past states and specifically addresses long-term dependency issues, making it perfectly suited for modeling a chaotic system. This is because the future state of the system predominantly relies on a limited number of preceding states.

\subsection{Reservoir Computing (RC)}
The most common realization of the RC approach are Echo state networks (ESNs), which have the advantage of low training cost which is superior in terms of low training cost comparing to other ML approaches. During the training process, all the parameters except the weights of output layer (readout layer) are randomly initialized and remain unchanged. Therefore, the weights of the output layer are adjusted and the training problem is turned into a linear regression problem.\par
\begin{figure}[t]
    \centering
    \includegraphics[width=1\linewidth]{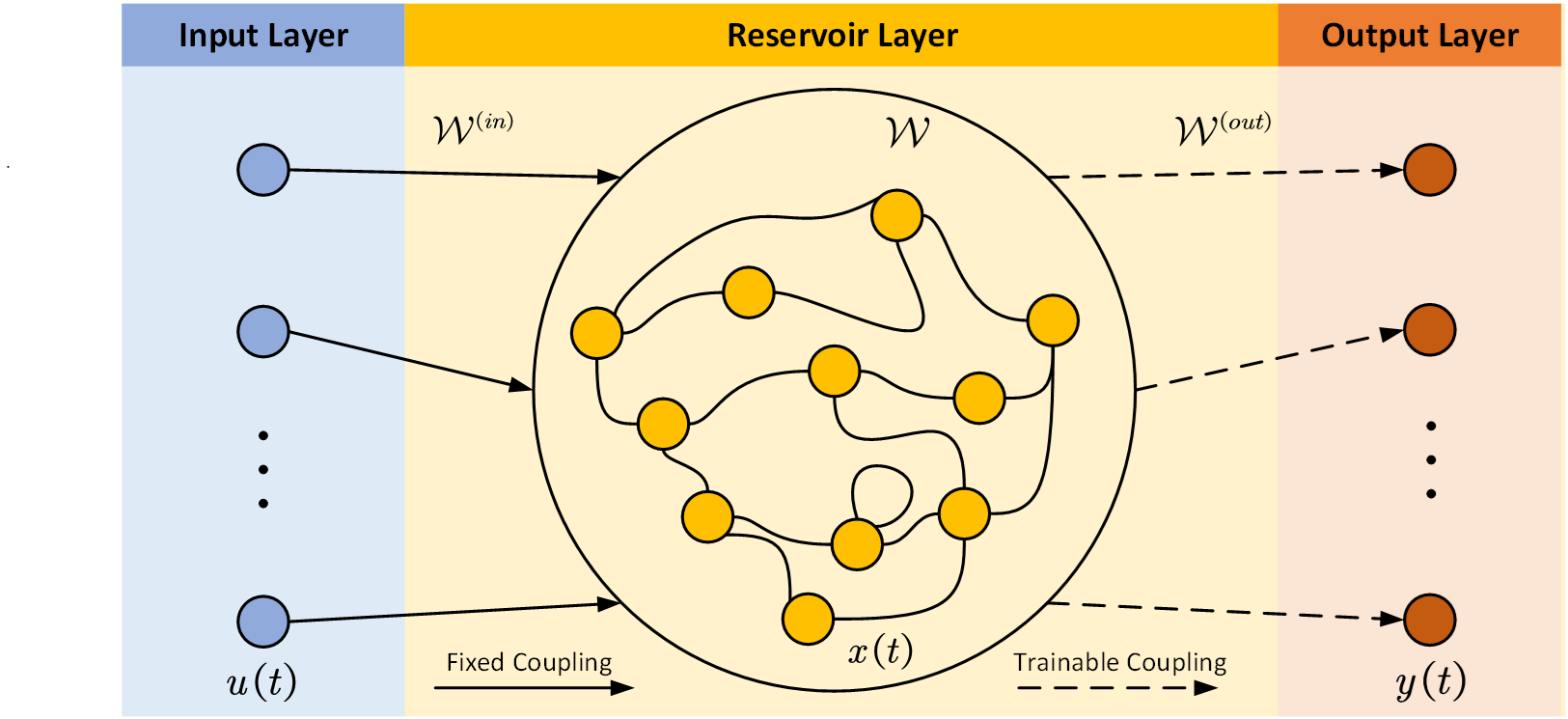}
    \caption{Schematic representation of the reservoir computing framework.}
    \label{fig:Reservoir Computing}
\end{figure}

RC is an effective method to model and predict nonlinear dynamics (e.g. chaotic dynamics). Fig. \ref{fig:Reservoir Computing} shows a typical type of RC, which consists of a input layer, a reservoir layer (also known as hidden layer) and a output layer. The input layer and output layer sizes are specified by nonlinear system's input and output sizes. The raw inputs are transformed and linked to the internal part of the RC, which consists of a coupled dynamical system with multiple nodes forming a recurrent network. The internal states of this system are calculated according to predetermined computational rules, and the outputs are generated by combining these internal states. In a standard RC setup, the connections from input to internal states and within the internal network are fixed, while only the connection from internal states to output is trained. This contrasts with general RNNs, where the entire network is trainable.
RC can generally be mathematically described as the following:
\begin{equation}
    \begin{aligned}
    &x(t+1) = (1-\gamma)x(t) + \gamma f(\mathcal{W}x(t)+\mathcal{W}^{(in)}u(t) + b)\\
    &y(t) = \mathcal{W}^{(out)} x(t)     
    \end{aligned}
\end{equation}
where $f$ is usually a component-wise nonlinear activation function (e.g. \(\tanh(\cdot)\)). And the input and output are mapped by weights $\mathcal{W}^{(in)}$ and $\mathcal{W}^{(out)}$. On the other hand, the internal network weight is $\mathcal{W}$. The constant parameters $b$ and $\gamma$ are introduced to ensure that the states $x(t)$ remain bounded, non-diminishing, and ideally exhibit rich patterns, facilitating subsequent extraction. The output weight is derived from minimizing a loss function. Assuming we now get some time series data ${z(t)}$ which is typically to be normalized for ensuring the model fairness and stability. Once the RC system is established by fixing the choice of all other params $f, \mathcal{W}^{(in)}, \mathcal{W}, \gamma$ and $b$ during the training process, the output weight can be obtained as, e.g.
\begin{equation}
    \mathcal{W}^{(out)} = \arg \underset{w}{\min}(||Xw-z||^2 + \beta ||w||^2)
\end{equation}
where~$x=[x(1)^{\top}, x(2)^{\top},\cdots, x(T)^{\top}]^{\top}$,~$z=[z(1)^{\top}, z(2)^{\top},\cdots, z(T)^{\top}]^{\top}$ and $\beta \in[0,1]$ is a penalization term which prevent $w$ from becoming too large. This problem is then formulated as Tikhonov regularization which yields an explicit solution $\mathcal{W}^{(out)\top} = (X^{\top}X+\beta I)^{-1}X^{\top}z$.\par

\subsection{Next Generation Reservoir Computing (NGRC)}
\begin{figure}[t]
    \centering
    \includegraphics[width=1\linewidth]{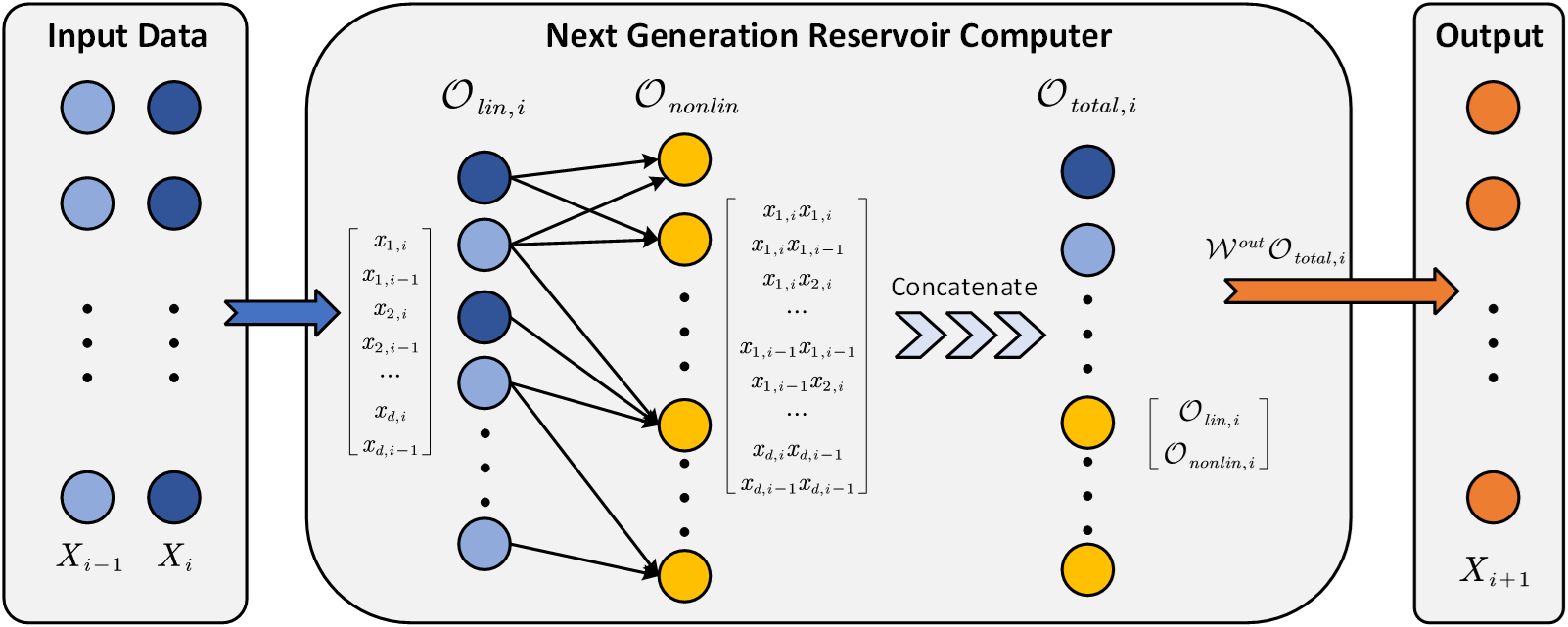}
    \caption{Schematic representation of the NGRC framework.}
    \label{fig:NGRC}
\end{figure}

Nonlinear vector auto regression (NVAR), also called next generation of reservoir computing (NG-RC), requires shorter training datasets and training time compared with RC \cite{gauthier2021next}. The NG-RC doesn't need a neural network, instead it uses a feature vector directly from the discretely sampled input data. As shown in Fig. \ref{fig:NGRC}, $\mathcal{O}_{total} = c \oplus 
\mathcal{O}_{lin} \oplus \mathcal{O}_{nonlin}$. where $\oplus$ represents the vector concatenation operation. $c$ is a constant, $\mathcal{O}_{lin}$, $\mathcal{O}_{nonlin}$ represent linear part and nonlinear part of feature vector, respectively. And the output weight matrix has the same form with RC.
\begin{equation}
    \mathcal{W}^{(out)} = z \ \mathcal{O}_{total}^{\top}(\mathcal{O}_{total}\mathcal{O}_{total}^{\top}+\beta I)^{-1}
\end{equation}
The linear features $\mathcal{O}_{lin, i}$ at time $i$ are made up of observations from input vector $X$, where $X_i=[x_{1,i}, x_{2,i}, \cdots, x_{d,i}]$, $d$ denotes its dimension. These observations are taken at current time steps and at $k-1$ previous time steps, with each time steps spaced by $s$. Here $s-1$ represents the number of skipped steps between consecutive observations. Thus $\mathcal{O}_{lin}$ has $dk$ components and is given by
\begin{equation}
    \mathcal{O}_{lin} = X_i \oplus X_{i-s} \oplus X_{i-2s} \oplus \cdots \oplus X_{i-(k-1)s}
\end{equation}
The nonlinear part $\mathcal{O}_{nonlin}$ of the feature vector is a nonlinear polynomial of $\mathcal{O}_{lin}$, where a $p^{\textrm{th}}$-order polynomial feature vector can be formulated as:
\begin{equation}
    \mathcal{O}_{nonlin} = \mathcal{O}_{lin} \otimes \mathcal{O}_{lin} \otimes \mathcal{O}_{lin} \cdots \otimes \mathcal{O}_{lin}
\end{equation}
with $\mathcal{O}_{lin}$ reoccurring $p$ times, where $\otimes$ denotes outer product which only collects the unique monomials in a vector.
Another aspect that NG-RC excels regular RC is that it requires a shorter warm-up period of $sk$ time steps, while traditional RC needs longer warm-up period to ensure that the reservoir state has the independence of the RC initial conditions. Reducing the warm-up time can be advantageous in scenarios where data is hard to obtain or when collecting more data is too time-consuming.

\section{Simulation and Results}
\subsection{Dataset Generation}
\begin{figure*}[t]
    \centering
    \includegraphics[width=1\linewidth]{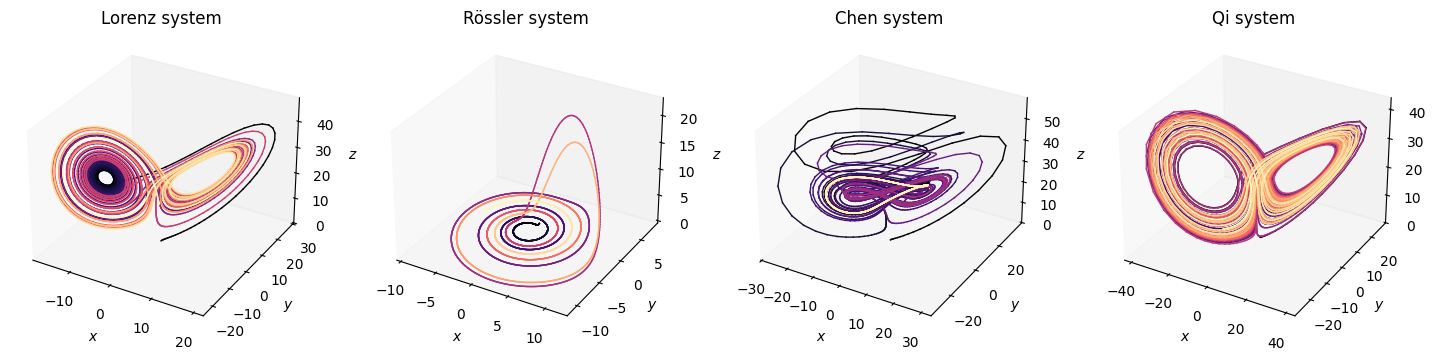}
    \caption{Four representative chaotic systems a) Lorenz System b) Rössler System c) Chen System d) Qi System }
    \label{4-sys}
\end{figure*}
In terms of getting real system dataset, we select four representative chaotic system \cite{lu2002bridge, qi2005analysis} shown in Fig. \ref{4-sys}, which consists of 5000 data points for each state with time step $dt = 0.01s$ using the RK23 method in Python.
\subsubsection{Lorenz System}
The following equation describes the Lorenz system:
\begin{equation}
\begin{aligned}
    &\dot{x} = \sigma(y-x) \\
    &\dot{y} = x(\rho-z)-y \\ 
    &\dot{z} = xy-\beta z \\
\end{aligned}
\end{equation}
where constants $\sigma, \rho, \beta$ are system parameters. The system exhibits chaotic behavior for these (and nearby) values $\sigma=10, \rho=28, \beta=\frac{8}{3}$.

\subsubsection{Rössler System}
The following equation describes the Rössler system:
\begin{equation}
\begin{aligned}
    &\dot{x} = -y-z \\
    &\dot{y} = x+ay \\ 
    &\dot{z} = (x-c)z+b \\
\end{aligned}
\end{equation}
we use the parameters $a=0.2, b=0.2, c=5.7$ for studying the chaotic behavior of Rössler System.

\subsubsection{Chen System}
Chen system is a double scroll chaotic attractor which defined as follows:
\begin{equation}
\begin{aligned}
    &\dot{x} = a(y-x) \\
    &\dot{y} = (c-a)x-xz+cy \\ 
    &\dot{z} = xy-bz \\
\end{aligned}
\end{equation}
we use the parameters $a=40, b=28, c=3$ for studying the chaotic behavior of Chen System.

\subsubsection{Qi System}
The following equation describes the Qi system:
\begin{equation}
\begin{aligned}
    &\dot{x} = a(y-x)+yz \\
    &\dot{y} = (a-c)x-y-xz \\ 
    &\dot{z} = xy-bz \\
\end{aligned}
\end{equation}
we use the parameters $a=35, b=7, c=10$ for studying the chaotic behavior of Qi System.


To evaluate the performance of these approaches in multi-step forecasting of chaotic time series, the aforementioned methods are applied to predict four chaotic benchmarks with the same initial condition $[x_0, y_0, z_0] = [17.6771581, 12.9313791, 43.9140433]$. 

\subsection{Task 1: Multi-steps forecasting of chaotic system behavior}
\begin{figure}[htpb]
    \centering
    \includegraphics[width=1\linewidth]{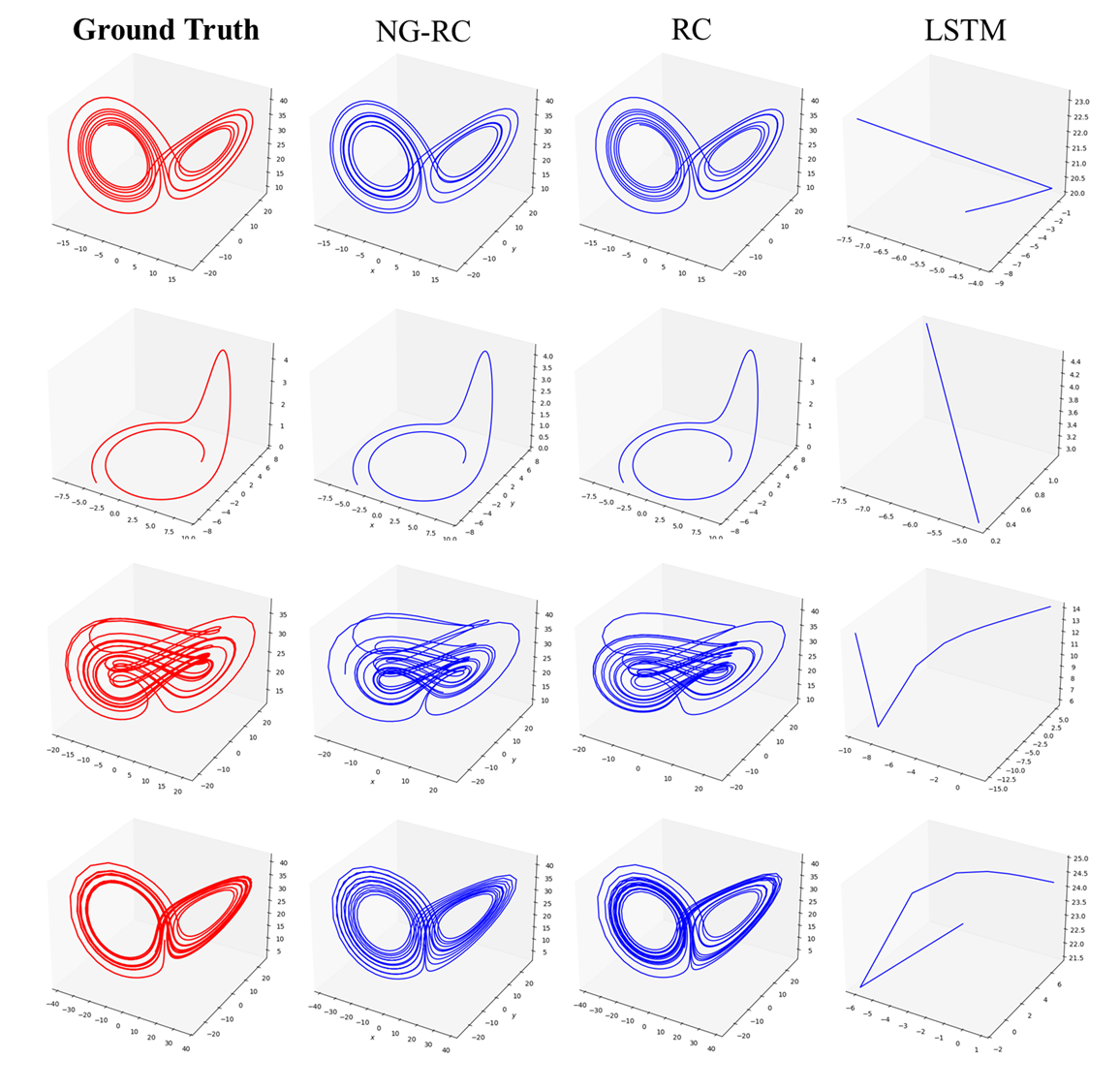}
    \caption{Comparation of 3 methods' prediction results}
    \label{task1_1}
\end{figure} 
For task 1, we divide the dataset into train, test dataset \footnote{Data used for warming-up RC and NG-RC are contained in train dataset} with ratio 8:2. For LSTM and RC approach, each dimension of dataset are scaled into $[0,1]$ using \texttt{minmaxscaler} to ensure training stability and accelerate training convergence. LSTM network's input and output dimension are set to 3 with a hidden layer of size 16 and are trained with 300 epochs. RC's parameters are set as below Table \ref{RC's param}.
\begin{table}[htpb]
\centering
\caption{Hyperparameters of RC}
\begin{tabular}{ll}
\hline
\textbf{Params} & \textbf{Value} \\ \hline
Number of Units & 4000 \\
Leak rate & 0.3 \\
Spectral radius & 1.25 \\
Connectivity & 0.1 \\
Regularization & 1e-8 \\ \hline
\label{RC's param}
\end{tabular}
\end{table}
NG-RC's params delay and strides are set to $k=2$ and $s=1$ with order $p=2$. Regularization term $\beta = 2.5\times 10^{-6}$.
After setting up all methods, we run it in Python on a single-core desktop processor. As shown in Fig. \ref{task1_1}, a qualitative comparison between the NG-RC, RC's predicted results and actual system reveals a high degree of similarity, indicating that NG-RC and RC can reproduce the long-term chaotic dynamics process with adequate training dataset. In the other side, the results of LSTM show that it can not reproduce the chaotic system dynamics even is trained well in single step forecasting. This make sense because in one step forecasting task, LSTM may perform well since it only needs to input past data. But in this task, LSTM network needs to use model output as new input to generate the entire chaotic behavior recursively. So we conclude the difference between LSTM and the other two methods are the latter learn the intrinsic of chaotic system while the former only learns the trend of data.\par
\begin{figure}[t]
    \centering
    \includegraphics[width=1\linewidth]{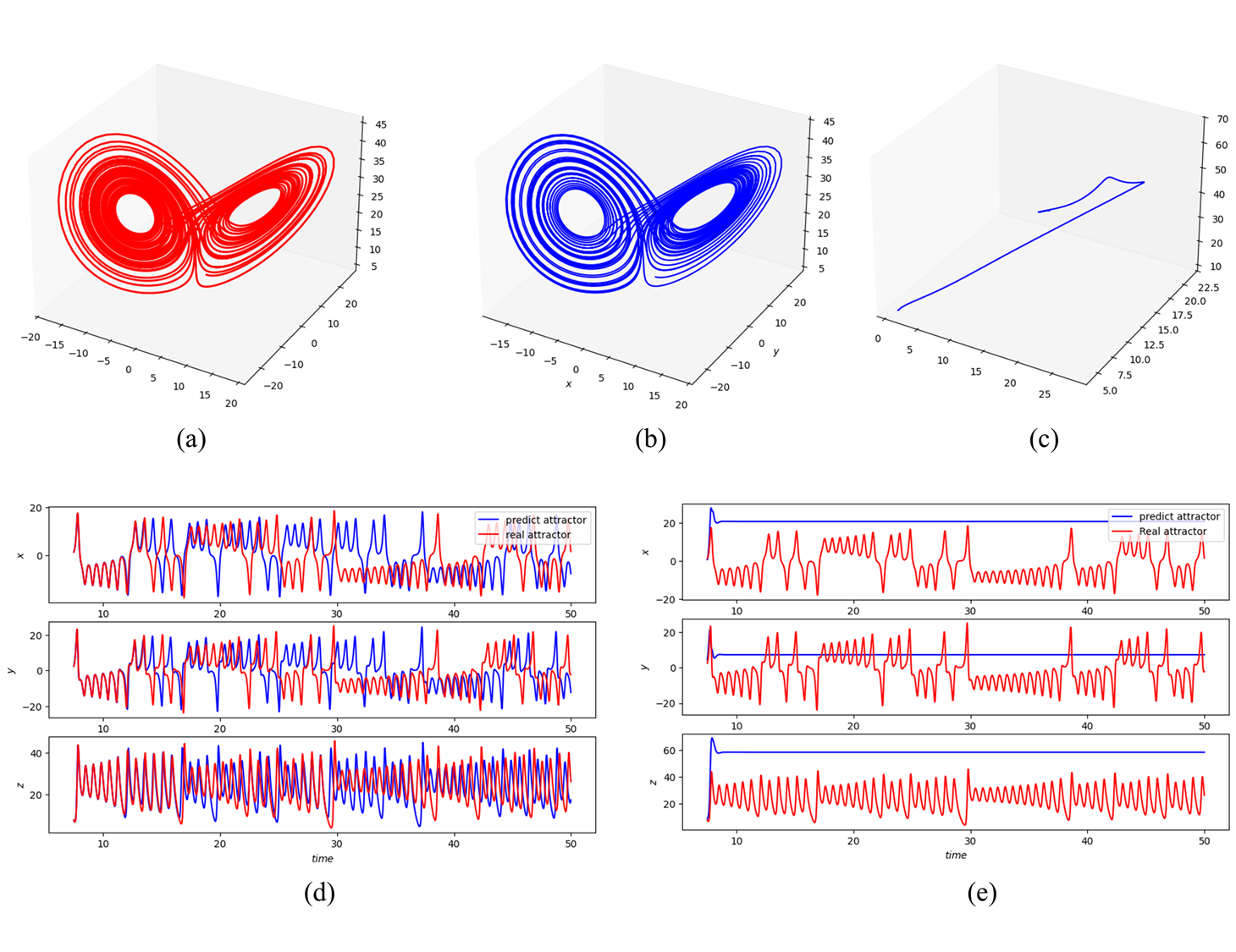}
    \caption{Comparison results of training RC and NG-RC with a smaller dataset. (a) Real attractor (b) NG-RC predict attractor (c) RC predict attractor (d) Each state of NG-RC predict attractor (blue) (e) Each state of RC predict attractor (blue)}
    \label{task1_2}
\end{figure}
\begin{figure}[t]
    \centering
    \includegraphics[width=1\linewidth]{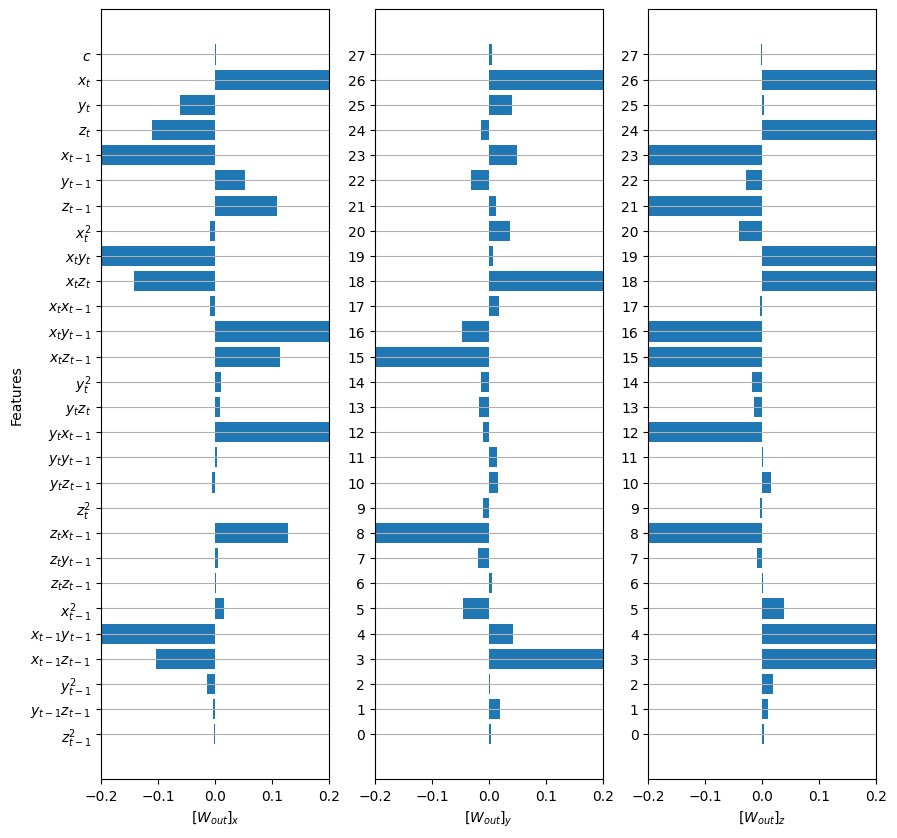}
    \caption{The output weights distribution of NG-RC in task 1}
    \label{weight}
\end{figure}

\begin{table}[h]
\centering
\caption{Training time comparation}
\begin{tabular}{llll}
\hline
Time & NG-RC & RC & LSTM \\ \hline
Lorenz & \textbf{0.46s} & 0.72s & 13.41s \\
Rössler & \textbf{0.49s} & 0.76s & 14.05s \\
Chen & \textbf{0.50s} & 0.84s & 14.37s \\
Qi & \textbf{0.46s} & 0.75s & 13.92s \\ \hline
\label{training_time}
\end{tabular}
\end{table}
Compared with RC, NG-RC shows almost the same predicting ability with RC in this task and only exceed in training time shown in Table \ref{training_time}. So we do another comparison experiment with RC and NG-RC which uses only a small training dataset to train and predicts the rest of dataset. In this experiment, the first 500 and 250 data points are used for training and warming-up, separately. And the rest 4250 data points are testing data points which used for comparing with generated attractor. RC and NG-RC's parameter setting are the same as above. As shown in Fig. \ref{task1_2},  RC can not predict Lorenz's chaotic behavior in the whole prediction length due to the lack of training dataset. However, it can be seen that the NG-RC can still perform well in small input training dataset scenario. Even there exists gap between real attractor and predicted attractor, they are similar in shape which indicates that NG-RC learns the topology of Lorenz's system and has a clear interpretability of its model. The output weights distribution of NG-RC readout layer are shown in Fig. \ref{weight}.\par

\begin{figure}[t]
    \centering
    \includegraphics[width=1\linewidth]{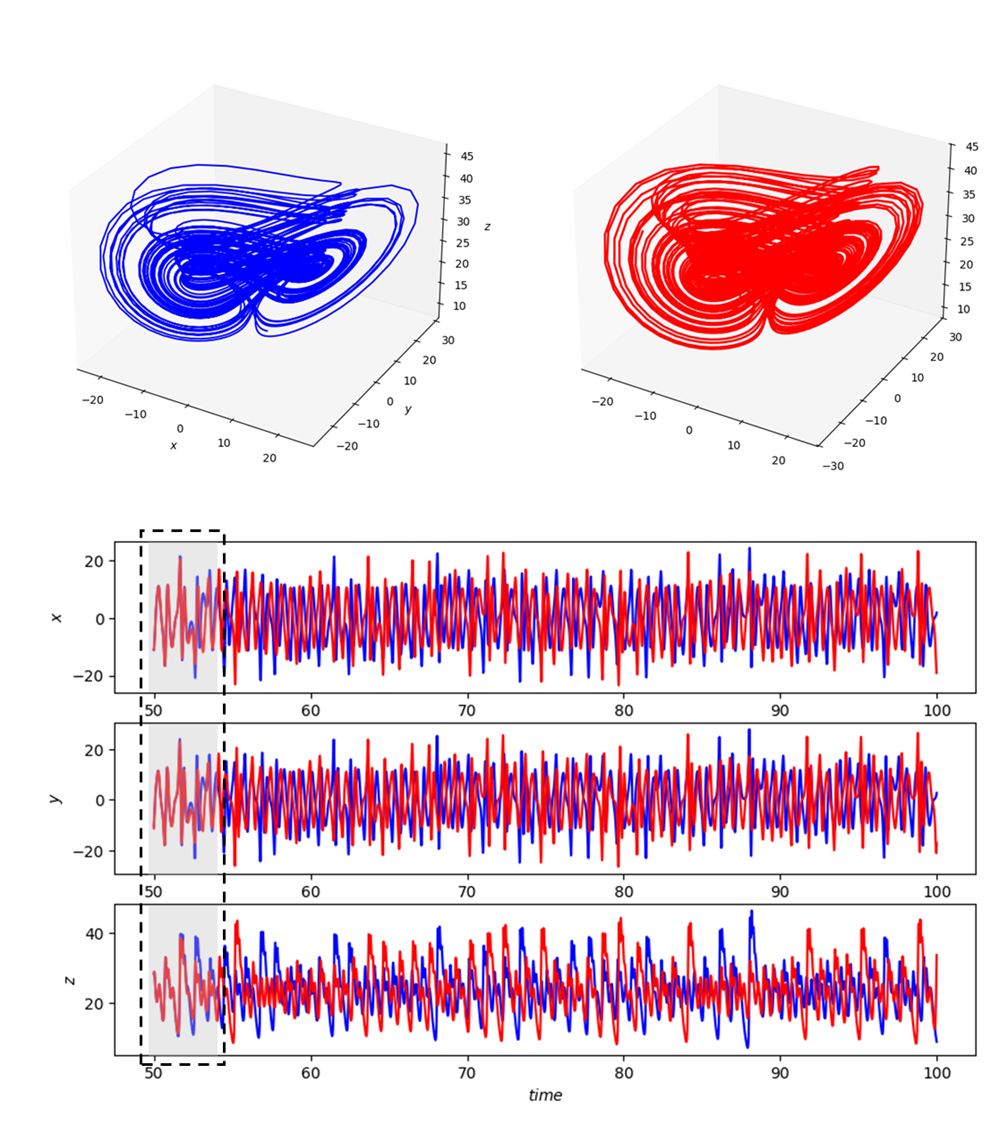}
    \caption{NG-RC predict Chen system in long time horizon. The prediction in the beginning of test time horizon (gray area with black dash line) exhibits good performance, which shows NG-RC's desirable forecasting ability.}
    \label{task1_3}
\end{figure}

Also, to better present NG-RC's better performance in relatively long time horizon, we also extend the dynamic system process time to 100s. In other words, we use a total 10000 data points to train and test NG-RC. In this setting we use 2000 and 3000 points to warm-up, training, respectively. And the rest 5000 points are used for testing. The results are shown in Fig. \ref{task1_3}.\par

It can be seen the prediction on the state vector deviates from the ground truth in the plots from each $x,y,z$ dimension, yet from the 3-D state trajectory plot, the prediction captures the overall system behavior adequately. Overall, NG-RC and RC outperforms in prediction with trained with sufficient training dataset comparing with LSTM. And NG-RC also show better prediction results in the experiment which RC and NG-RC trained only with a small dataset, which show data efficiency of NG-RC. It is worth mentioning that NG-RC also has an explicit interpretability comparing to other methods and show good prediction results in long time horizon prediction task.\par

\subsection{Task 2: Predicting System State With Noisy Input Data}
\begin{figure*}[t]
    \centering
    \includegraphics[width=1\linewidth]{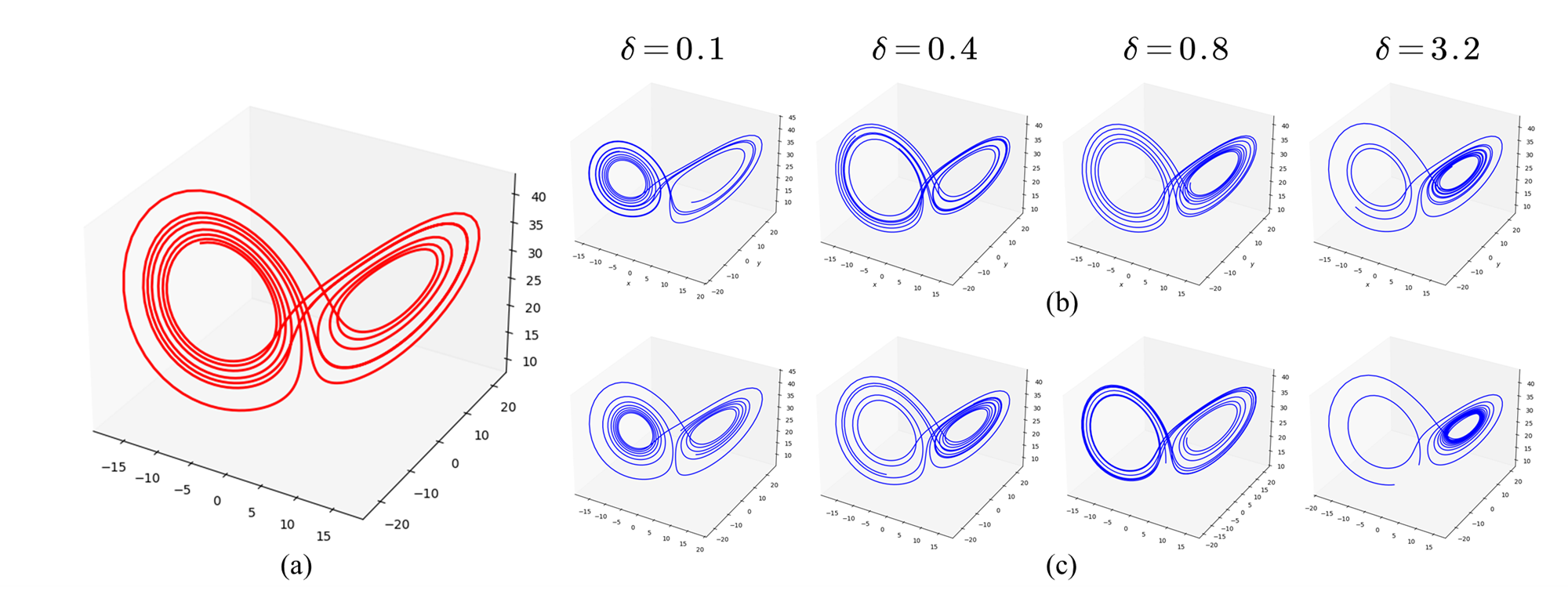}
    \caption{Prediction performance of noisy Lorenz system for RC and NG-RC with different noise level. (a) Original real attractor (b) NG-RC predicting results (c) RC predicting results}
    \label{task2}
\end{figure*}
To benchmark NG-RC and RC's robustness on noisy systems, we use the following noisy Lorenz system to generate a dataset with process noise.
\begin{equation}
    \begin{aligned}
    &\dot{x} = 10(y-x)+\sigma\mathcal{E}_1 \\
    &\dot{y} = x(28-z)-y+\sigma\mathcal{E}_2 \\ 
    &\dot{z} = xy-\frac{8}{3}z+\sigma\mathcal{E}_3 \\
    \end{aligned}
\end{equation}
where Gaussian noise $\mathcal{E}_i \in \mathcal{N}(0,1), i=1,2,3$. $\sigma$ is the magnitude of noise. Except for these modifications, RC and NG-RC's parameters are set as the same as task 1. 3750 and 250 data points are used for training and warming-up. Prediction results of RC and NG-RC given in Fig. \ref{task2} shows well performance encountering different noise level. Even for Lorenz with high level noise, both RC and NG-RC show similar predicting results that reproduce the Lorenz System. It indicates RC and NG-RC has nearly the same robustness with noise. More experiments with different noise level are presented in Appendix.

\section{Conclusion}
In this paper, three current popular ML time-series predicting approaches including NG-RC, RC and LSTM were tested to forecast four representative chaotic time series which are generated by Lorenz, Rössler, Chen, Qi system.  
In the task of predicting chaotic time series data (Task 1), the NG-RC method outperforms other methods in all aspects due to its fast computational speed and fewer adjustable parameters. It not only has a light-weight network structure but also provides interpretability to the model, which reveals the linear and nonlinear components weights and henceforth the impact of each weight to the model. Furthermore, we have shown that the reduced size of training datasets causes RC to be unable to predict chaotic time series well, while it has no impact on the performance of NG-RC. Also, as the dataset's dimension increase, the number of adjustable parameters of RC and LSTM will dramatically explode. 


In addition, we also tested the NG-RC's performance in the long time horizon and it gets satisfactory results in reproducing the entire chaotic system behavior. Although the generated chaotic attractor is not totally the same as real attractor, NG-RC still learn the topology of the real attractor. This also demonstrate NG-RC's ability to reproduce chaotic system behavior using only a small amount of observable data points.\par

Moreover, the robustness test of RC and NG-RC was presented. We added different level random Gaussian noise to each state of dynamical system during process and obtain a polluted time series dataset. RC is highly sensitive to noisy chaotic time series. We concluded that RC and NG-RC are both robust when encountering noisy chaotic system. Both of them can still reproduce the chaotic system even for chaotic system with high level noise.\par

In conclusion, using RC generally lacks interpretability \cite{shahi2022prediction}, as it functions like a black-box model with several hyperparameters that to be adjusted. While a trained RC model can effectively represent time series, it does not provide clear insights into the mathematical structure underlying the observed data. We observed NG-RC is best for learning dynamical systems due to its 1) network interpretability, 2) fewer parameters to optimize, 3) extremely less dataset for training. NG-RC can also be applied in various engineering scenarios, such as creating digital twins for dynamic systems using observed data or by integrating approximate models with observations for data assimilation. Additionally, when it is challenging to derive a nonlinear system using traditional methods, NG-RC offers a more effective solution compared to other data-driven approaches. Our future work will focuses on find an efficient way to predict chaotic system with different kind of noise and benchmark ML methods' ability on reconstructing unseen data using existing partially observable data, which is meaningful in the field of controls systems.\par

\bibliography{ref}

\begin{thebibliography}{10}

\bibitem{farmer1987predicting}
J.~D. Farmer and J.~J. Sidorowich, ``Predicting chaotic time series,'' {\em Physical review letters}, vol.~59, no.~8, p.~845, 1987.

\bibitem{muller1997predicting}
K.-R. M{\"u}ller, A.~J. Smola, G.~R{\"a}tsch, B.~Sch{\"o}lkopf, J.~Kohlmorgen, and V.~Vapnik, ``Predicting time series with support vector machines,'' in {\em International conference on artificial neural networks}, pp.~999--1004, Springer, 1997.

\bibitem{sapankevych2009time}
N.~I. Sapankevych and R.~Sankar, ``Time series prediction using support vector machines: a survey,'' {\em IEEE computational intelligence magazine}, vol.~4, no.~2, pp.~24--38, 2009.

\bibitem{weigend2018time}
A.~S. Weigend, {\em Time series prediction: forecasting the future and understanding the past}.
\newblock Routledge, 2018.

\bibitem{torres2021deep}
J.~F. Torres, D.~Hadjout, A.~Sebaa, F.~Mart{\'\i}nez-{\'A}lvarez, and A.~Troncoso, ``Deep learning for time series forecasting: a survey,'' {\em Big Data}, vol.~9, no.~1, pp.~3--21, 2021.

\bibitem{ahmed2010empirical}
N.~K. Ahmed, A.~F. Atiya, N.~E. Gayar, and H.~El-Shishiny, ``An empirical comparison of machine learning models for time series forecasting,'' {\em Econometric reviews}, vol.~29, no.~5-6, pp.~594--621, 2010.

\bibitem{shi2017deep}
H.~Shi, M.~Xu, and R.~Li, ``Deep learning for household load forecasting—a novel pooling deep rnn,'' {\em IEEE Transactions on Smart Grid}, vol.~9, no.~5, pp.~5271--5280, 2017.

\bibitem{tsai2018air}
Y.-T. Tsai, Y.-R. Zeng, and Y.-S. Chang, ``Air pollution forecasting using rnn with lstm,'' in {\em 2018 IEEE 16th Intl Conf on Dependable, Autonomic and Secure Computing, 16th Intl Conf on Pervasive Intelligence and Computing, 4th Intl Conf on Big Data Intelligence and Computing and Cyber Science and Technology Congress (DASC/PiCom/DataCom/CyberSciTech)}, pp.~1074--1079, IEEE, 2018.

\bibitem{mikhaeil2022difficulty}
J.~Mikhaeil, Z.~Monfared, and D.~Durstewitz, ``On the difficulty of learning chaotic dynamics with rnns,'' {\em Advances in Neural Information Processing Systems}, vol.~35, pp.~11297--11312, 2022.

\bibitem{hochreiter1997long}
S.~Hochreiter and J.~Schmidhuber, ``Long short-term memory,'' {\em Neural computation}, vol.~9, no.~8, pp.~1735--1780, 1997.

\bibitem{RC1}
J.~Pathak, B.~Hunt, M.~Girvan, Z.~Lu, and E.~Ott, ``Model-free prediction of large spatiotemporally chaotic systems from data: A reservoir computing approach,'' {\em Physical review letters}, vol.~120, no.~2, p.~024102, 2018.

\bibitem{RC2}
M.~Rafayelyan, J.~Dong, Y.~Tan, F.~Krzakala, and S.~Gigan, ``Large-scale optical reservoir computing for spatiotemporal chaotic systems prediction,'' {\em Physical Review X}, vol.~10, no.~4, p.~041037, 2020.

\bibitem{RC3}
G.~Carleo, I.~Cirac, K.~Cranmer, L.~Daudet, M.~Schuld, N.~Tishby, L.~Vogt-Maranto, and L.~Zdeborov{\'a}, ``Machine learning and the physical sciences,'' {\em Reviews of Modern Physics}, vol.~91, no.~4, p.~045002, 2019.

\bibitem{yan2024emerging}
M.~Yan, C.~Huang, P.~Bienstman, P.~Tino, W.~Lin, and J.~Sun, ``Emerging opportunities and challenges for the future of reservoir computing,'' {\em Nature Communications}, vol.~15, no.~1, p.~2056, 2024.

\bibitem{bo2020asynchronously}
Y.-C. Bo, P.~Wang, and X.~Zhang, ``An asynchronously deep reservoir computing for predicting chaotic time series,'' {\em Applied Soft Computing}, vol.~95, p.~106530, 2020.

\bibitem{griffith2019forecasting}
A.~Griffith, A.~Pomerance, and D.~J. Gauthier, ``Forecasting chaotic systems with very low connectivity reservoir computers,'' {\em Chaos: An Interdisciplinary Journal of Nonlinear Science}, vol.~29, no.~12, 2019.

\bibitem{pathak2018model}
J.~Pathak, B.~Hunt, M.~Girvan, Z.~Lu, and E.~Ott, ``Model-free prediction of large spatiotemporally chaotic systems from data: A reservoir computing approach,'' {\em Physical review letters}, vol.~120, no.~2, p.~024102, 2018.

\bibitem{gauthier2021next}
D.~J. Gauthier, E.~Bollt, A.~Griffith, and W.~A. Barbosa, ``Next generation reservoir computing,'' {\em Nature communications}, vol.~12, no.~1, pp.~1--8, 2021.

\bibitem{lu2002bridge}
J.~L{\"u}, G.~Chen, D.~Cheng, and S.~Celikovsky, ``Bridge the gap between the lorenz system and the chen system,'' {\em International Journal of Bifurcation and Chaos}, vol.~12, no.~12, pp.~2917--2926, 2002.

\bibitem{qi2005analysis}
G.~Qi, G.~Chen, S.~Du, Z.~Chen, and Z.~Yuan, ``Analysis of a new chaotic system,'' {\em Physica A: Statistical Mechanics and its Applications}, vol.~352, no.~2-4, pp.~295--308, 2005.

\bibitem{shahi2022prediction}
S.~Shahi, F.~H. Fenton, and E.~M. Cherry, ``Prediction of chaotic time series using recurrent neural networks and reservoir computing techniques: A comparative study,'' {\em Machine learning with applications}, vol.~8, p.~100300, 2022.

\end{thebibliography}
\bibliographystyle{ieeetr}
\newpage

\section*{Appendix}
\begin{figure}[h]
    \centering
    \includegraphics[width=0.8\linewidth]{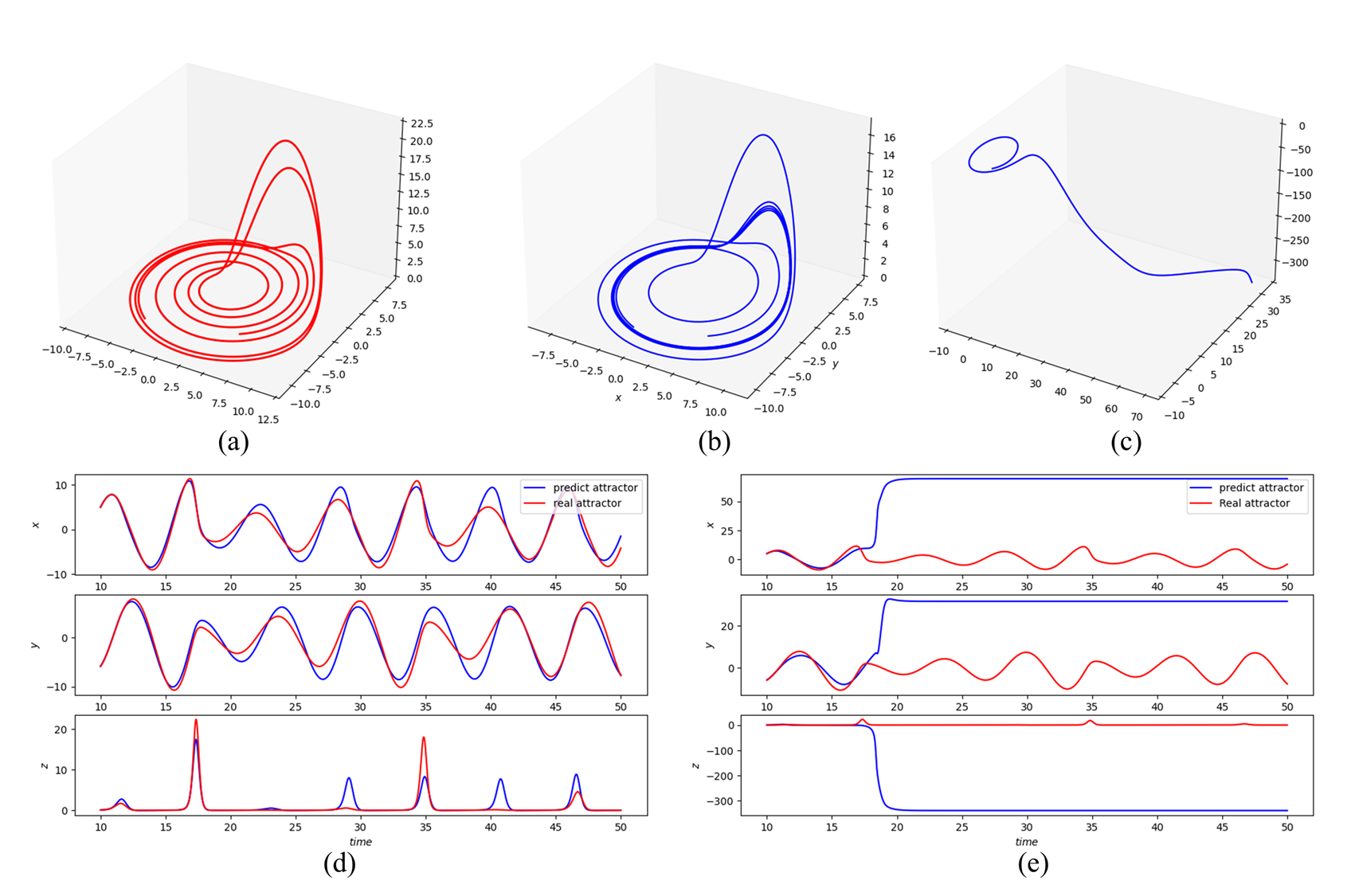}
    \caption{Comparison results of training RC and NG-RC with a smaller dataset of Rossler System (750 data points for training and 250 data points for warming-up). (a) Real attractor (b) NG-RC predict attractor (c) RC predict attractor (d) Each state of NG-RC predict attractor (blue) (e) Each state of RC predict attractor (blue)}
\end{figure}

\begin{figure}[h]
    \centering
    \includegraphics[width=0.8\linewidth]{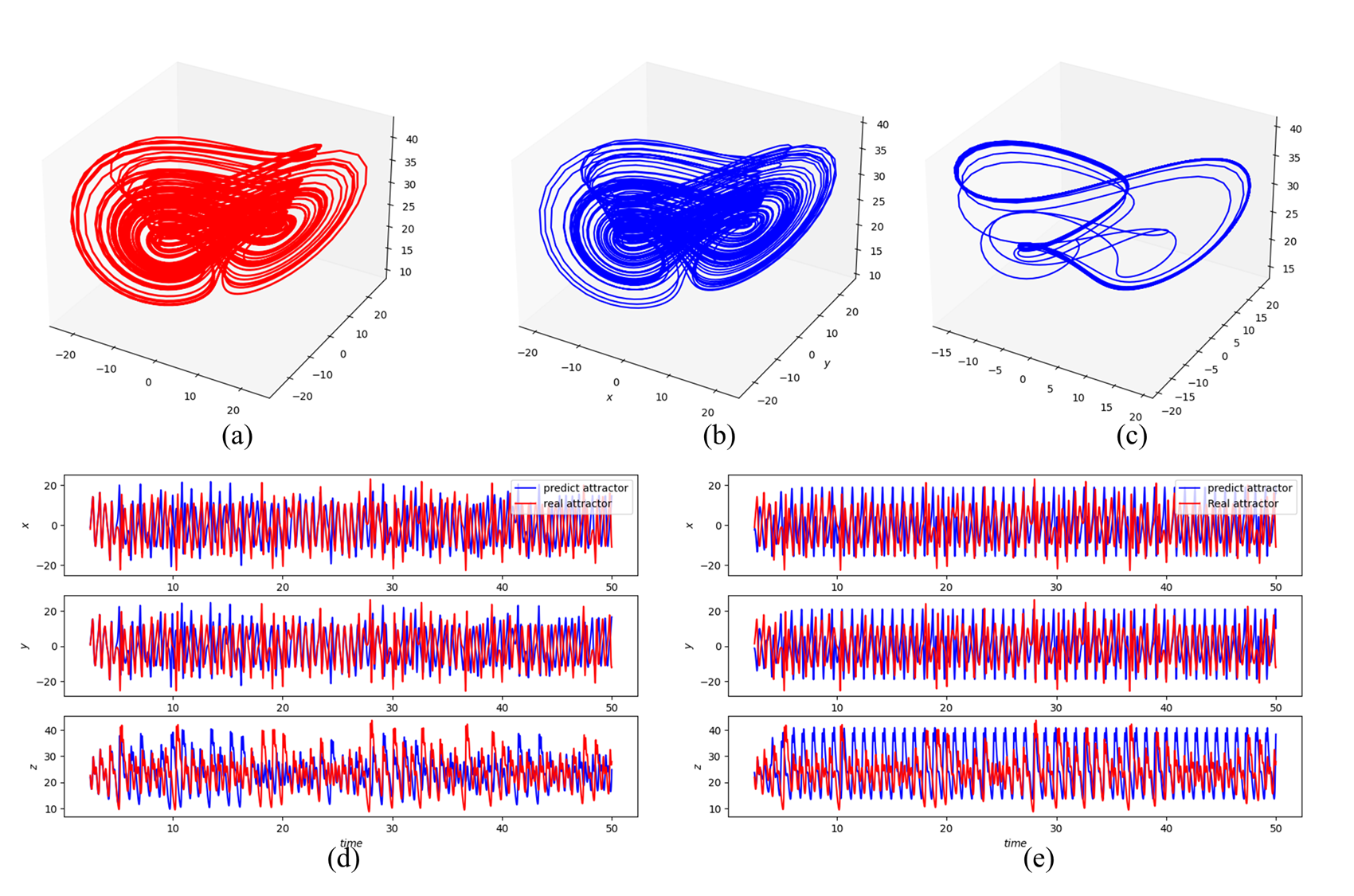}
    \caption{Comparison results of training RC and NG-RC with a smaller dataset of Chen System (150 data points for training and 100 data points for warming-up). (a) Real attractor (b) NG-RC predict attractor (c) RC predict attractor (d) Each state of NG-RC predict attractor (blue) (e) Each state of RC predict attractor (blue)}
\end{figure}

\begin{figure}[!h]
    \centering
    \includegraphics[width=0.8\linewidth]{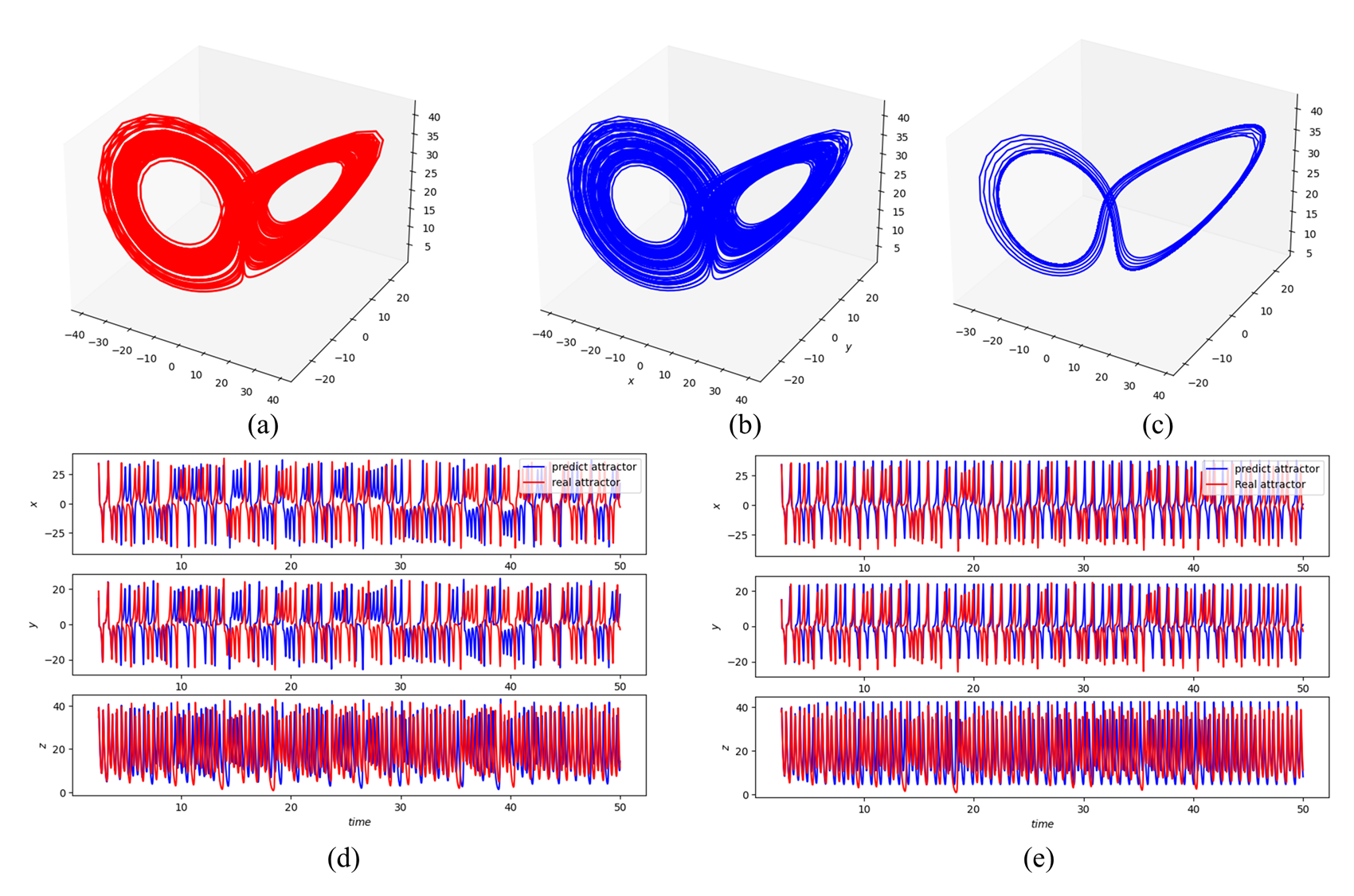}
    \caption{Comparison results of training RC and NG-RC with a smaller dataset of Qi System (150 data points for training and 100 data points for warming-up). (a) Real attractor (b) NG-RC predict attractor (c) RC predict attractor (d) Each state of NG-RC predict attractor (blue) (e) Each state of RC predict attractor (blue)}
\end{figure}

\newpage

\begin{figure*}[h]
    \centering
    \includegraphics[width=0.9\linewidth]{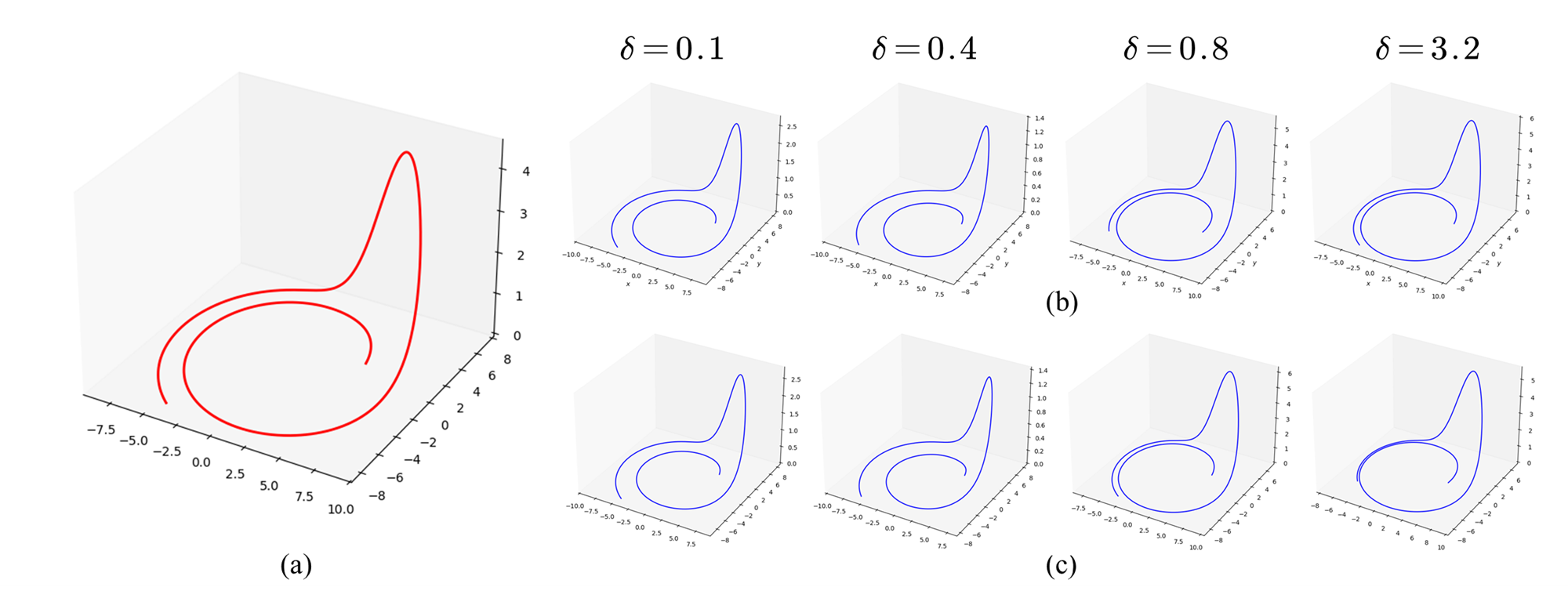}
    \caption{Prediction performance of noisy Rössler system for RC and NG-RC with different noise level. (a) Original real attractor (b) NG-RC predicting results (c) RC predicting results}
\end{figure*}

\begin{figure*}[h]
    \centering
    \includegraphics[width=0.9\linewidth]{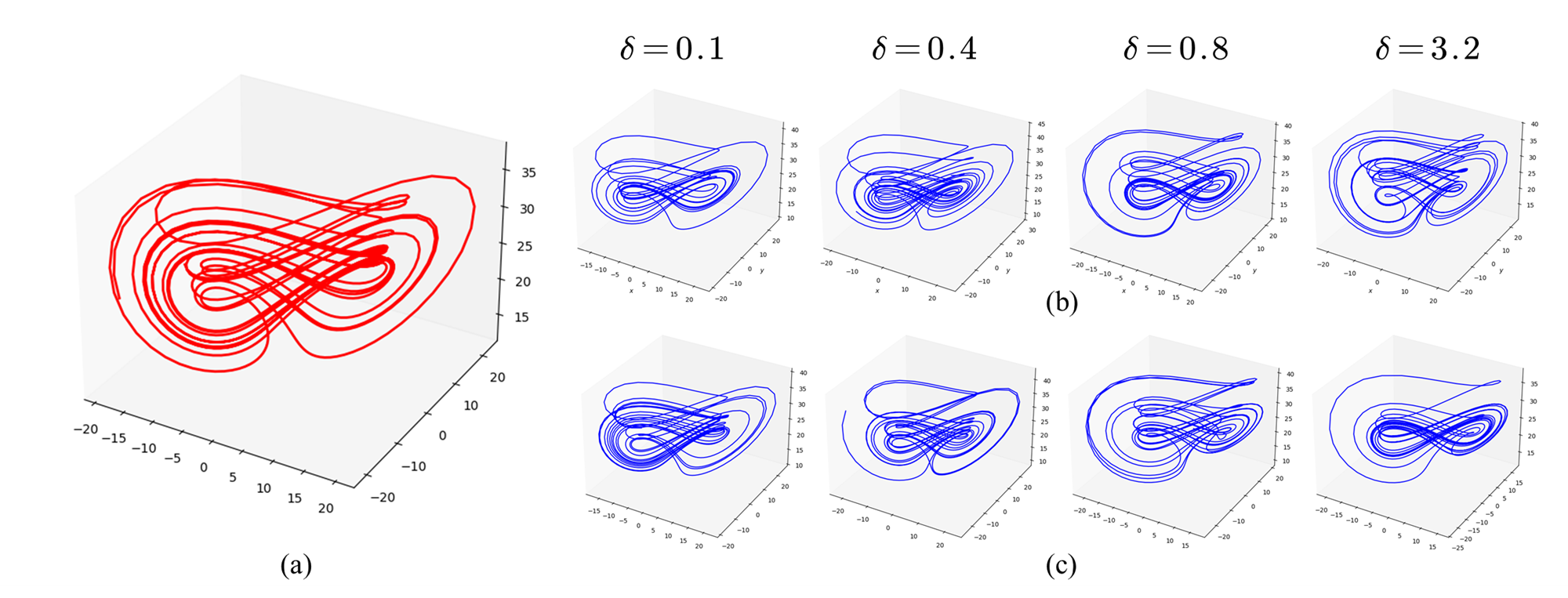}
    \caption{Prediction performance of noisy Chen system for RC and NG-RC with different noise level. (a) Original real attractor (b) NG-RC predicting results (c) RC predicting results}
\end{figure*}

\begin{figure*}[h]
    \centering
    \includegraphics[width=0.9\linewidth]{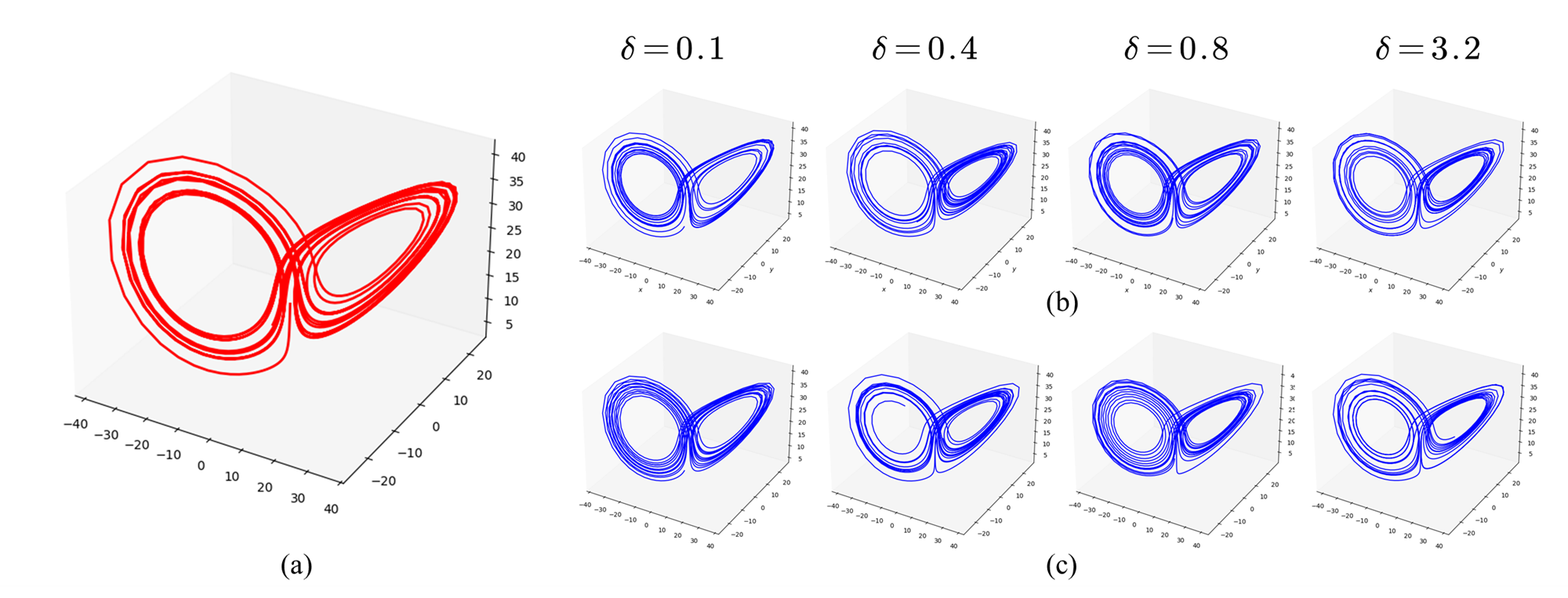}
    \caption{Prediction performance of noisy Qi system for RC and NG-RC with different noise level. (a) Original real attractor (b) NG-RC predicting results (c) RC predicting results}
\end{figure*}
\end{document}